%% file: iclr2027_conference.tex
\documentclass{article} 
\usepackage{iclr2027_conference,times}

\input{math_commands.tex}

\usepackage{hyperref}
\usepackage{url}
\usepackage{lipsum}
\usepackage{enumitem}

\usepackage{xspace}
\usepackage{graphicx}
\usepackage{booktabs}
\usepackage{array}
\usepackage{multirow} 
\usepackage{subcaption}

\usepackage{xspace}
\makeatletter
\DeclareRobustCommand\onedot{\futurelet\@let@token\@onedot}
\def\@onedot{\ifx\@let@token.\else.\null\fi\xspace}
\def\eg{\emph{e.g}\onedot} 
\def\ie{\emph{i.e}\onedot} 
\def\cf{\emph{c.f}\onedot} 

\makeatother

\newcommand{\methodname}{EgoJEPA\xspace}

\title{EgoJEPA: Beyond Retargeting for Dexterous Manipulation from Egocentric Videos}

\author{Meng Cao\textsuperscript{1}$^{*}$, ~Can Zhao\textsuperscript{2,3}\thanks{\ \ Equal contributions.}, ~Jinjie Wei\textsuperscript{1}$^{*}$, ~Haonan Zhao\textsuperscript{2,4}, ~Yulong Zhang\textsuperscript{3}, ~Duo Han\textsuperscript{3}, \\ \textbf{Xiaofan Li\textsuperscript{3}}, \textbf{Ziwu Song\textsuperscript{3}}, \textbf{Rongtao Xu\textsuperscript{1}}, \textbf{Ruyi Gan\textsuperscript{3}}, \textbf{Hao Wang\textsuperscript{3}}, \textbf{Qian Wang\textsuperscript{3}}, \\\textbf{Daolin Ma\textsuperscript{2,4}}$^{\dagger}$, \textbf{Ian Reid\textsuperscript{1}}, \textbf{Xiaodan Liang\textsuperscript{1,5}}\thanks{\ \ Corresponding author.}\\
\textsuperscript{1}Mohamed bin Zayed University of Artificial Intelligence \\
\textsuperscript{2}Shanghai Jiao Tong University
\textsuperscript{3}X Square Robot
\textsuperscript{4}Xense Robotics
\textsuperscript{5}Sun Yat-sen University\\
\url{https://mengcaopku.github.io/EgoJEPA}
}

\iclrfinalcopy 
\begin{document}

\maketitle

\vspace{-5mm}
\input{sec/0_abs}
\input{sec/1_intro}
\input{sec/2_related}

\input{sec/3_method}
\input{sec/4_exp}

\input{sec/5_con}
\input{sec/6_statement}

\bibliography{iclr2027_conference}
\bibliographystyle{iclr2027_conference}
\clearpage
\appendix
\input{sec/6_appendix}

\end{document}

%% file: math_commands.tex
\usepackage{amsmath,amsfonts,bm}

\def\eqref#1{equation~\ref{#1}}

\def\1{\bm{1}}

\DeclareMathAlphabet{\mathsfit}{\encodingdefault}{\sfdefault}{m}{sl}
\SetMathAlphabet{\mathsfit}{bold}{\encodingdefault}{\sfdefault}{bx}{n}



%% file: sec/0_abs.tex
\begin{abstract}
Egocentric human videos offer a scalable alternative to costly robot demonstrations, yet transferring their manipulation priors to robot control remains challenging. The commonly adopted retargeting-based approaches convert human wrist and finger poses into embodiment-specific action commands through hand-crafted geometric mappings. However, such mappings require meticulous recalibration across different embodiments and risk propagating hand-tracking errors into downstream robot control pipelines. To this end, we propose \textbf{EgoJEPA}, a three-stage framework that learns embodiment-agnostic representations from egocentric human videos without rule-based human-to-robot retargeting: 1) \emph{Human-action pre-training} learns effect-aware latent actions through JEPA-style visual prediction conditioned on hand-motion differences; 2) \emph{Latent-action-based world modeling} jointly predicts future visual dynamics and latent actions, which decouples human prior pre-training from robot-specific control parameterizations; 3) \emph{Embodiment-specific tuning} adapts the pretrained model to diverse robots through lightweight action adapters. Experiments on five real-world manipulation tasks across three robot embodiments demonstrate the effectiveness and efficiency of EgoJEPA. It achieves a 74\% average success rate under the out-of-domain evaluation, compared with 65\% for the retargeting baseline, while reducing cumulative cross-embodiment wall-clock cost by 47.2\%, from 743.2 to 392.7 hours.
\end{abstract}

%% file: sec/1_intro.tex
\section{Introduction} \label{sec:intro}
\vspace{-1mm}
Despite recent advances in robotic manipulation policies \citep{chi2025diffusion,zhao2023learning,kim2024openvla}, scaling such systems remains constrained by their reliance on task-specific robot demonstrations, typically obtained through teleoperation or specialized capture systems \citep{zhao2023learning,khazatsky2024droid,chi2024universal,sun2026maniparena,zhang2026wall}. Collecting these demonstrations is costly, labor-intensive, and time-consuming, posing a significant bottleneck to scalable robot learning. In contrast, \emph{egocentric human videos} provide a scalable supervision source for transferring manipulation skills from human hand-object interactions to robotic systems~\citep{ma2026robot,feng2026human}.

Earlier \emph{action-free} methods operate without explicit hand-pose annotations, instead learning latent action representations directly from inter-frame visual dynamics \citep{ye2025latent,wang2026unified,bjorck2025gr00t,gao2026dreamdojo,bharadhwaj2024gen2act}. However, the resulting representations may capture generic visual variations rather than action-induced physical effects, limiting their interpretability and transferability to robot control. To move beyond learning solely from visual supervision, recent approaches explicitly leverage human hand-pose annotations and adopt a \emph{retargeting-based pipeline} \citep{kareer2025egomimic,yang2025egovla,tao2025dexwild,ma2026humanscale,wang2026humanego}. Specifically, human hand motions, such as wrist and finger poses, are captured using specialized devices (\eg, Leap Motion \citep{weichert2013analysis} or Aria glasses \citep{engel2023project}) and subsequently mapped to a target robot action space through hand-crafted rules for policy pre-training or human-robot co-training. 

\input{table_figs/figTeaser}


Despite the intuitive formulation, the retargeting pipeline suffers from two key limitations: 1) \emph{Embodiment-specific mapping}. Retargeting relies on low-level geometric or kinematic mappings tailored to a particular robot embodiment, rather than learning transferable action semantics. Consequently, such mappings are tightly coupled to the robot morphology and typically require meticulous recalibration whenever either changes. As shown in Figure \ref{fig:trainingTimeSplit} (a), human--robot calibration alone takes 52.1 hours, accounting for 25.2\% of the total pipeline time. Figure~\ref{fig:trainingTimeSplit} (b) shows that this cost compounds when scaling to multiple robots, \ie, retargeting must repeat the embodiment-specific human-robot calibration and pre-training whenever the robot morphology changes, which causes the cumulative wall-clock time to grow rapidly. 2) \emph{Noise propagation}. Retargeting is highly sensitive to errors in estimated human motion. Hand-pose estimation noise caused by occlusions, viewpoint variations, or temporal jitter can propagate through the retargeting process, resulting in physically infeasible robot actions. As a concrete stress test, we inject Gaussian noise into human action annotations while keeping robot training unchanged. As shown in Figure~\ref{fig:robustness} (a), under a mild per-axis position noise of $\sigma=10$~mm, the retargeting pipeline on the task of \texttt{bottle cap unscrewing} drops drastically from 60\% to 30\% success rate, showing that annotation errors are directly amplified into downstream robot control failures.

To this end, we propose \textbf{EgoJEPA}, an \textbf{Ego}centric instantiation of the \textbf{J}oint \textbf{E}mbedding \textbf{P}redictive \textbf{A}rchitecture \citep{assran2023self} that learns embodiment-agnostic latent action representations from human videos. As shown in Figure \ref{fig:trainingTimeSplit}, EgoJEPA comprises three stages: 1) \emph{Human-action pre-training} learns generic effect-aware latent actions through human-motion-conditioned visual prediction; 2) \emph{Latent-action-based world modeling} jointly predicts future visual dynamics and the \emph{latent} actions learned in Stage~1 via joint flow matching, rather than embodiment-specific robot commands; and 3) \emph{Embodiment-specific tuning} adapts the pre-trained world model to robot action spaces through lightweight adapters. Evaluated on five real-world manipulation tasks across three robot embodiments (two dexterous hands and a gripper), EgoJEPA achieves the highest average success rate of 74\% under the out-of-domain evaluation, outperforming the action-free method LAPA \citep{ye2025latent}, the retargeting pipeline, and the prior egocentric human-video method DexWild \citep{tao2025dexwild}. In addition, EgoJEPA significantly reduces the cost of scaling across robot embodiments. By reusing Stage~1--2 models and adapting only Stage~3 for each new robot, EgoJEPA reduces cumulative wall-clock cost from 743.2 to 392.7 hours over retargeting (\cf Figure~\ref{fig:trainingTimeSplit} (b)).



Our contributions are summarized as follows:
\begin{itemize}[topsep=0pt, partopsep=0pt, leftmargin=13pt, parsep=0pt, itemsep=3pt]
    \item We propose EgoJEPA, a JEPA-style framework that eliminates the need for rule-based human-to-robot retargeting by learning embodiment-agnostic representations from human videos. 

    \item We introduce effect-aware latent actions as a transferable interface between human motion priors and robot control, enabling efficient policy adaptation across diverse embodiments via lightweight action adapters.       

    \item Extensive experiments on five real-world manipulation tasks across three robot embodiments demonstrate the effectiveness and cross-embodiment scalability of EgoJEPA. 
\end{itemize}

%% file: table_figs/figTeaser.tex
\begin{figure*}[t]
	\centering
    \includegraphics[width=0.98\textwidth]{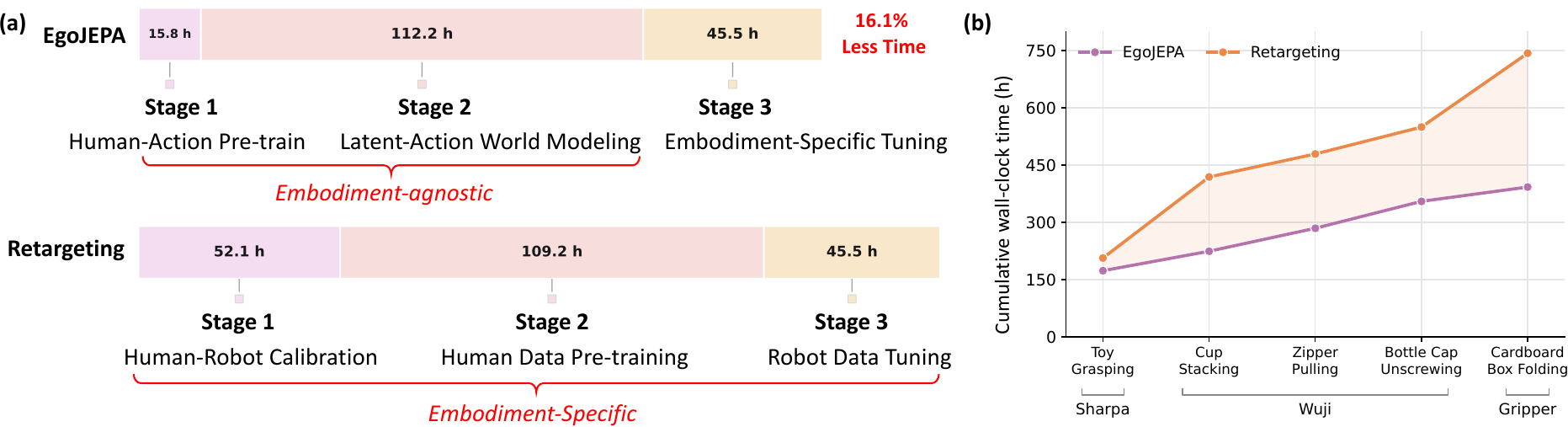}
    \vspace{-1mm}
    \caption{\textbf{Training efficiency and scalability of EgoJEPA \emph{v.s.} retargeting.} \textbf{(a)} The retargeting pipeline requires embodiment-specific human-to-robot mappings and repeated calibration, whereas EgoJEPA learns an embodiment-agnostic latent action space in Stages~1--2 and performs embodiment-specific grounding only in Stage~3; \textbf{(b)} Cumulative training cost across robot embodiments. Reusing the embodiment-agnostic models from Stages~1--2 reduces the per-embodiment adaptation overhead of EgoJEPA relative to the retargeting pipeline.}
	\label{fig:trainingTimeSplit}
\end{figure*}

%% file: sec/2_related.tex
\section{Related Work}\label{sec:related}

\noindent \textbf{WAMs for Dexterous Manipulation.} WAMs move beyond action-only policies by incorporating future state prediction into the action learning process, thereby modeling the coupled evolution of visual dynamics and action control \citep{wang2026world,shen2026world,li2025comprehensive}. 
While early WAMs mainly target low-DoF gripper manipulation, recent works have extended this paradigm toward dexterous manipulation \citep{goswami2025world,kim2026dexterous,jiang2409dexsim2real2}. DexWM \citep{goswami2025world} conditions latent-state prediction on fine-grained hand keypoint displacements for dexterous interaction modeling. DWM \citep{kim2026dexterous} simulates dexterous behaviors using scene-action-conditioned video diffusion without explicitly decoding robot actions. However, existing dexterous WAMs rely on embodiment-specific robot demonstrations, limiting cross-embodiment transfer and scalability. EgoJEPA instead learns transferable manipulation priors from egocentric human videos, enabling efficient adaptation to diverse robots.


\noindent \textbf{Robot Learning from Human Videos.} Due to the high cost of teleoperated robot demonstrations, recent work has increasingly turned to human-centric robot learning, which leverages abundant human videos as scalable semantic and physical priors \citep{ma2026robot,feng2026human}. Existing approaches can be broadly grouped into two categories. 1) \emph{Latent-based methods} learn action-related representations from \emph{action-free} videos by encoding inter-frame changes \citep{ye2025latent,wang2026unified,bjorck2025gr00t,gao2026dreamdojo} or future visual dynamics \citep{wu2024unleashing,cheang2024gr,bharadhwaj2024gen2act}. However, lacking explicit action grounding, such latents may conflate controllable hand motion with camera motion and passive scene dynamics, making them less interpretable as action representations. 2) \emph{Retargeting-based methods} recover explicit human motion, \eg, wrist poses and finger poses captured by devices like Leap Motion or Aria glasses, and map them to robot-specific action spaces for pre-training or co-training \citep{kareer2025egomimic,yang2025egovla,tao2025dexwild,ma2026humanscale,wang2026humanego,zheng2026egoscale}. 
Although these approaches provide more structured supervision, they remain tied to embodiment-specific geometric mappings and are sensitive to hand-tracking noise. To this end, EgoJEPA encodes explicit human hand motions into effect-aware latent actions, enabling transfer to robot control without rule-based retargeting while improving robustness to hand-tracking noise. While concurrent work such as Being-H0 \citep{luo2025being} adopts a similar spirit by discretizing MANO trajectories into motion tokens via a VQ-VAE structure, its motion tokens rely exclusively on isolated hand-pose sequences without grounding the tokenizer in the visual context. Consequently, such tokenizers are agnostic to visual interaction semantics, failing to capture the critical contact dynamics and spatial alignment between hands and target objects.



\noindent \textbf{Joint-Embedding Predictive Architectures.} JEPAs formulate latent world modeling by predicting the future evolution of a system in a compact representation space rather than reconstructing raw observations \citep{lecun2022path}. Early works apply JEPA to self-supervised representation learning, where models learn semantic representations by predicting latent embeddings of masked image or video patches \citep{assran2023self,bardes2024v,assran2025v}. Recent studies further explore JEPA for action-conditioned latent world modeling \citep{assran2025v,zhou2024dino}. LeWorldModel \citep{maes2026leworldmodel} extends JEPA into an end-to-end action-conditioned world model via SIGReg regularization \citep{bardes2021vicreg}. DINO-WM \citep{zhou2024dino} instead leverages pretrained visual representations to learn latent dynamics by predicting future semantic features. Following this spirit, EgoJEPA investigates how human motion can serve as a physical condition for latent world modeling, which enables adaptation to diverse robot embodiments.








%% file: sec/3_method.tex
\section{EgoJEPA}\label{sec:method}
\input{table_figs/figPipeline}


We propose \methodname{}, a framework that learns embodiment-agnostic action representations from egocentric human videos and efficiently adapts them to downstream robotic manipulation tasks. As illustrated in Figure~\ref{fig:pipeline}, \methodname{} comprises three key stages: human-action pre-training (Stage~1), latent-action-based world modeling (Stage~2), and embodiment-specific tuning (Stage~3).

\subsection{Stage 1: Human-Action Pre-training} \label{sec:stage1}

Instead of relying on rule-based, embodiment-specific retargeting pipelines, \methodname{} first learns an effect-aware latent space grounded in \emph{human-action}-conditioned visual prediction. Specifically, we define the ``human action'' as the difference between adjacent-frame hand annotations, such as 2D/3D keypoints, MANO parameters \citep{romero2022embodied}, or other hand-pose representations. This difference reflects the local hand motion responsible for visual state transitions, providing an embodiment-agnostic physical latent which can be later adapted to different robot embodiments.

Let $\mathcal{D}_{\mathrm{ego}}$ denote an egocentric human video dataset, where each trajectory contains RGB observations $\{o_t\}_{t=1}^{T}$ and aligned hand annotations $\{m_t\}_{t=1}^{T}$. We instantiate the \emph{human action} over a horizon $H$ as adjacent-frame hand annotation differences.
\begin{equation}
    \Delta m_{t:t+H-1} = \big(m_{i+1}-m_i\big)_{i=t}^{t+H-1}.
\end{equation}
We then optimize a human-action-conditioned JEPA objective \citep{assran2023self} that predicts the target visual representation from visual context and hand-motion differences, thereby grounding the learned action-effect latent in visual state transitions.
\begin{equation}
    \operatorname*{minimize}_{\phi,\psi,\theta}
    \left\|
        \boldsymbol{P}_{\phi}
        \Big(
            \big(\boldsymbol{E}_{\theta}(o_i)\big)_{i=t}^{t+H-1},
            \boldsymbol{A}_{\psi}(\Delta m_{t:t+H-1})
        \Big)
        -
        \mathrm{sg}
        \big(
            \boldsymbol{E}_{\overline{\theta}}(o_{t+H})
        \big)
    \right\|_1 ,
\end{equation}
where $\boldsymbol{E}_{\theta}$ and $\boldsymbol{E}_{\bar{\theta}}$ denote the online and target visual encoders, respectively, with $\bar{\theta}$ updated by an exponential moving average of $\theta$. The action encoder $\boldsymbol{A}_{\psi}$ maps the human action chunk $\Delta m_{t:t+H-1}$ into the latent action embedding. The predictor $\boldsymbol{P}_{\phi}$ then combines this action condition with the context representations $(\boldsymbol{E}_{\theta}(o_i))_{i=t}^{t+H-1}$ to predict the target representation at time $t+H$. The stop-gradient operator $\mathrm{sg}(\cdot)$ prevents gradients from flowing through the target branch.

\subsection{Stage 2: Latent-Action-Based World Modeling} \label{sec:stage2}

Building on Stage~1, we train a \emph{latent-action world model} to jointly predict future visual dynamics and the corresponding latent actions. Notably, the action branch is supervised by frozen JEPA action embeddings $\boldsymbol{A}_{\psi}(\Delta m)$, avoiding the reliance on retargeted coordinate-level robot actions. 

Given an egocentric trajectory, we first construct the \emph{clean} targets for the future visual dynamics and the latent action sequences as follows.
\begin{equation}
    z_1 = \boldsymbol{E}_{\theta}(o_{t+1:t+H}),
    \qquad
    u_1 = \boldsymbol{A}_{\psi}(\Delta m_{t:t+H-1}),
\end{equation}
where $z_1$ represents the visual dynamics, while $u_1$ denotes the corresponding latent action embedding. $\boldsymbol{E}_{\theta}$ and $\boldsymbol{A}_{\psi}$ denote the visual and action encoders adopted from the Stage~1 pre-trained model, respectively. We then formulate latent-action world modeling as a joint flow matching problem over future visual representations and latent actions. Specifically, for a flow timestep $\tau \in [0,1]$, we construct the noisy joint state by linear interpolation:
\begin{equation}
    z_\tau = \tau z_1 + (1-\tau)z_0,
    \qquad
    u_\tau = \tau u_1 + (1-\tau)u_0,
\end{equation}
where $z_0 \sim \mathcal{N}(0,I)$ and $u_0 \sim \mathcal{N}(0,I)$ are Gaussian noises with the same dimensionalities as $z_1$ and $u_1$. The world-action model $\boldsymbol{W}_{\omega}$ is trained to predict the velocity field from the noisy joint state to the clean video-action target $[z_1,u_1]$:
\begin{equation}
\mathcal{L}_{\mathrm{Stage2}}(\omega)
=
\mathbb{E}_{z,u,\tau}
\left[
\left\|
\boldsymbol{W}_{\omega}
\left(
[z_\tau,u_\tau],
q,
\tau
\right)
-
\left(
[z_1,u_1]-[z_0,u_0]
\right)
\right\|_2^2
\right],
\end{equation}
where $q$ denotes the language instruction, and $[z_\tau,u_\tau]$ denotes the noisy output token sequence at flow timestep $\tau$, consisting of future visual tokens $z_\tau$ and latent action tokens $u_\tau$. 

\subsection{Stage 3: Embodiment-Specific Tuning} \label{sec:stage3}

In Stage~3, EgoJEPA grounds the shared action representation into a target robot embodiment. Specifically, we initialize the world-action model $\boldsymbol{W}_{\omega}$ from Stage~2 and insert lightweight embodiment-specific adapters into its action branch. As shown in Figure \ref{fig:pipeline} (right), the \emph{input adapter} maps robot action tokens into the shared action-token space of the pretrained WAM, while the \emph{output adapter} maps the WAM action-channel output back to the target robot action space.


To align the robot supervision with the relative hand-motion representations learned in Stage~1, we express robot actions as temporal differences. Given a sequence of absolute robot states $\{a_i\}_{i=t}^{t+H}$, the delta-action chunk is defined as follows.
\begin{equation}
    \Delta a_{t:t+H-1}
    =
    \left\{a_{i+1}-a_i\right\}_{i=t}^{t+H-1},
    \label{eq:delta_robot_action}
\end{equation}
where $H$ denotes the action prediction horizon. Let $o_t$ denote the visual observation at time $t$, and let $o_{t+1:t+H}$ denote the subsequent RGB observation sequence over a horizon of $H$. We construct the clean future visual target and robot delta-action target as
\begin{equation}
    z_1 = \boldsymbol{E}_{\theta}(o_{t+1:t+H}),
    \qquad
    \Delta a_1 = \Delta a_{t:t+H-1},
\end{equation}
where $z_1$ denotes the future visual dynamics of the robot trajectory, and $\Delta a_1$ denotes the corresponding executable relative robot action chunk. 
We then construct the noisy visual and action states through linear interpolation:
\begin{equation}
    z_\tau = \tau z_1 + (1-\tau)z_0,
    \qquad
    \Delta a_\tau = \tau \Delta a_1 + (1-\tau)\Delta a_0,
\end{equation}
where $z_0 \sim \mathcal{N}(0,I)$ and $\Delta a_0 \sim \mathcal{N}(0,I)$ are the sampled Gaussian noises. 

Let $\boldsymbol{W}_{\omega}$ denote the pretrained world-action model from Stage~2, and let $\boldsymbol{D}_{\eta}$ denote the action adapter for the specific robot embodiment. The Stage~3 objective is formulated as follows.
\begin{equation}
\mathcal{L}_{\mathrm{Stage3}}(\omega,\eta)
=
\mathbb{E}_{z,\Delta a,\tau}
\left[
\left\|
\boldsymbol{D}_{\eta} \circ \boldsymbol{W}_{\omega}
\left(
[z_\tau,\Delta a_\tau],
q,
\tau
\right)
-
\left(
[z_1,\Delta a_1]-[z_0,\Delta a_0]
\right)
\right\|_2^2
\right],
\end{equation}
where $q$ denotes the language instruction, and $[z_\tau,\Delta a_\tau]$ denotes the noisy output token sequence composed of future visual tokens and robot delta-action tokens. We use $\boldsymbol{D}_{\eta}^{r} \circ \boldsymbol{W}_{\omega}$ to denote the adapted world-action model, where the embodiment-specific adapter is applied to the action channel while the visual encoder $\boldsymbol{E}_{\theta}$ is kept unchanged. 

%% file: table_figs/figPipeline.tex
\begin{figure*}[t]
	\centering
        \includegraphics[width=\textwidth]{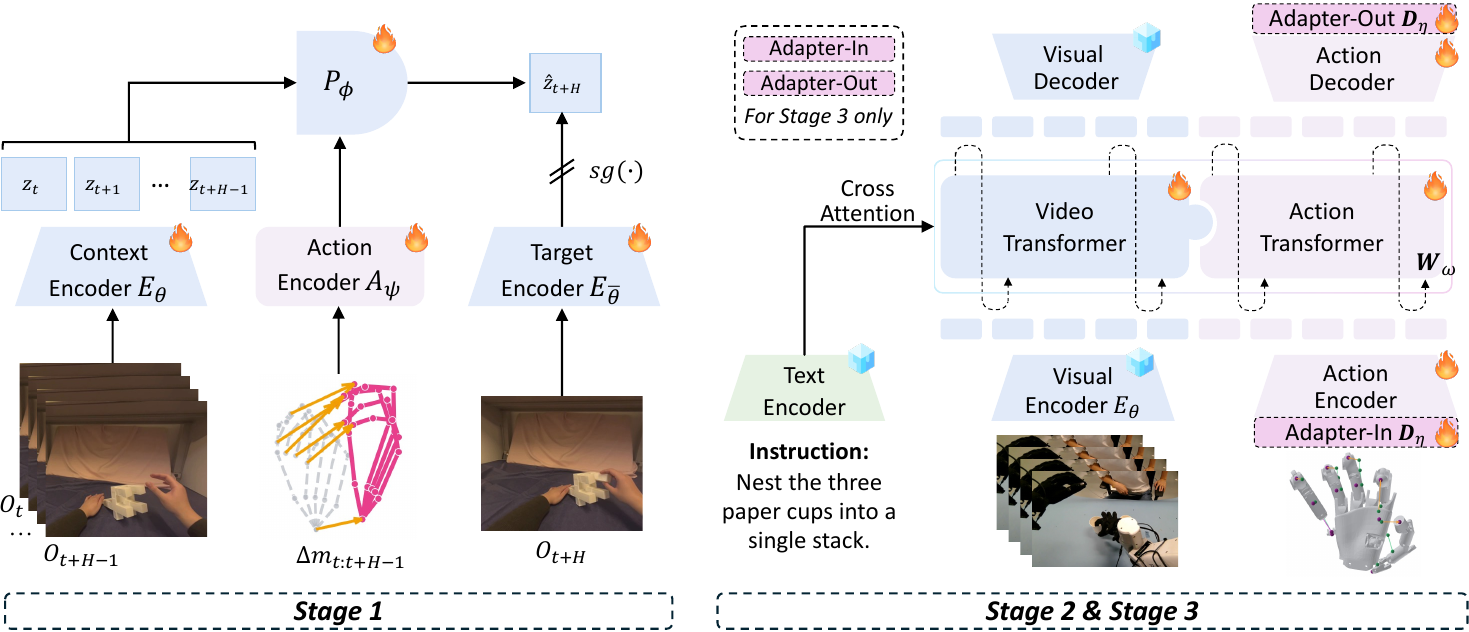}
        \caption{\textbf{Overview of the EgoJEPA framework.} \textbf{Stage~1}: An effect-aware latent action space is learned from ego-centric human video data without embodiment-specific action re-targeting. Specifically, the predictor $\boldsymbol{P}_{\phi}$ predicts the future representation $\hat{z}_{t+H}$ from the visual context encoded by $\boldsymbol{E}_{\theta}$ and the hand-motion differences $\Delta m_{t:t+H-1}$ encoded by $\boldsymbol{A}_{\psi}$; \textbf{Stage~2}: A world-action model $\boldsymbol{W}_{\omega}$ is trained on ego-centric human video data via joint flow matching to model future visual dynamics and latent human actions extracted by the frozen action encoder $\boldsymbol{A}_{\psi}$; \textbf{Stage~3}: Using robot demonstration data, the learned action representation is grounded in a target robot embodiment by inserting embodiment-specific input and output adapters $\boldsymbol{D}_{\eta_r}$ into the action branch, which map between relative robot actions and the shared latent action space.}
	\label{fig:pipeline}
\end{figure*}

%% file: sec/4_exp.tex
\section{Experiments}\label{sec:exp}


\subsection{Experimental Setup}

\noindent\textbf{Human Pre-training Data.} We use a commercially acquired egocentric human dataset for stages~1 and~2. The dataset contains 1,111,943 episodes from 62,518 recording sessions, amounting to approximately 20,106.6 hours and 2.17 billion video frames. It covers 64,686 task instructions and 608 scene labels across diverse household and industrial activities. Each episode consists of a monocular RGB video recorded at approximately 30Hz, together with camera calibration and motion annotations comprising wrist poses and 21 three-dimensional keypoints for each hand.

\noindent \textbf{Robot Post-training Tasks.} We evaluate EgoJEPA on five real-world manipulation tasks spanning three robot embodiments including a Sharpa dexterous hand \citep{sharpa_wave}, a Wuji dexterous hand \citep{wuji_hand}, and a Xense Tac UMI gripper \citep{xense_xtac_umi}. Each task is associated with a task-specific robot demonstration dataset for Stage~3 post-training: 1) \emph{Toy grasping} uses the Sharpa hand to grasp a plush toy; 2) \emph{Cup stacking} uses the Wuji hand to arrange cups into a target stack, requiring sequential manipulation and precise placement; 3) \emph{Zipper pulling} uses the Wuji hand to pinch and slide a zipper pull; 4) \emph{Bottle cap unscrewing} uses the Wuji hand to uncap a bottle through coordinated grasping; 5) \emph{Cardboard box folding} uses the  Xense Tac UMI gripper to fold a paper box into a target configuration, requiring multi-step manipulation. See Appendix~\ref{app:datasets} for detailed specifications on both human pre-training and robot post-training datasets.

\noindent\textbf{Implementation Details.} In Stage~1, the context and target visual encoders are initialized from the pretrained Wan2.2 VAE encoder~\citep{wan2025}, and the predictor is a 24-layer Transformer with frame-causal attention. In Stage~2, the video Transformer is initialized from Wan2.2-TI2V-5B~\citep{wan2025}, while the Stage~1 action embeddings $u_1$ are average-pooled along the feature dimension to obtain 128-dimensional latent action targets for stable training. In Stage~3, the input and output action adapters each comprise a two-layer MLP with a ReLU activation between the layers. Full training configurations are provided in Table~\ref{tab:training_config}. 


\noindent\textbf{Evaluation Protocol.} We evaluate all policies under both in-domain and out-of-domain settings across five tasks and three robot embodiments. In-domain evaluation uses the same object instances and spatial layouts as Stage~3 training. Out-of-domain evaluation considers two forms of distribution shift: 1) \emph{object shifts}, using unseen object instances with different shapes, sizes, or materials while preserving task requirements; 2) \emph{spatial shifts}, placing objects at positions and orientations outside the Stage~3 training distribution but within the robot's reachable workspace. All policies are evaluated without additional fine-tuning under matched test configurations. We conduct 20 trials per task for each setting and define success as full task completion.

\noindent \textbf{Compared Methods.} We evaluate \methodname{} against five competitive methods and architectural variants: 1) \emph{LAPA} \citep{ye2025latent}, an action-free method that learns discrete latent actions directly from inter-frame visual changes for policy pre-training; 2) \emph{DexWild} \citep{tao2025dexwild}, an egocentric human-video method that co-trains dexterous manipulation policies on in-the-wild videos and robot demonstrations; 3) \emph{Retargeting-based pipeline}, which shares our architecture but replaces the learned latent actions with embodiment-specific robot targets derived via geometric retargeting, followed by fine-tuning on robot data; 4) \emph{EgoJEPA (Robot-only)}, an ablation that omits Stage~1--2 human-video pre-training and is trained strictly on robot demonstrations; and 5) \emph{EgoJEPA (VQ-VAE)}, a variant that replaces Stage~1 JEPA pre-training with a VQ-VAE reconstruction objective \citep{van2017neural} over hand-motion annotations, while maintaining identical Stage~2--3 protocols. Refer to Appendix~\ref{app:implementation} for more implementation details.

\subsection{Real-World Performance}

\input{table_figs/figMainRes}
\input{table_figs/figVisMainResults}

We present in-domain and out-of-domain manipulation results in Figure~\ref{fig:mainRes}, highlighting three key findings: 1) \emph{State-of-the-art performance.} EgoJEPA attains the highest success rate on all five tasks, with average success rates of 94\% and 74\% under in- and out-of-domain settings, respectively, surpassing the retargeting pipeline by 14\% and 9\% absolute margins. 2) \emph{Generalization gap.} All methods exhibit lower average success rates under out-of-domain evaluation, highlighting the challenges posed by unseen object instances and spatial layouts. Nevertheless, EgoJEPA achieves the highest out-of-domain success rate of 74\%, compared with 65\% for retargeting and 40\% for DexWild. 3) \emph{Task-level variation.} \texttt{Bottle cap unscrewing} and \texttt{cardboard box folding} yield EgoJEPA's lowest out-of-domain success rates. The former requires coordinated bimanual control to stabilize and rotate the bottle, while the latter involves multi-step operations with strict geometric alignments. These constraints increase sensitivity to contact noise and error accumulation, making reliable completion challenging.

\subsection{Ablation Studies}

We conduct ablation studies to evaluate EgoJEPA's architectural design and data efficiency. All policy evaluations follow the out-of-domain setting.

\noindent\textbf{Benefits of Human Pre-training.} To assess the contribution of human pre-training to downstream manipulation, we compare EgoJEPA with a robot-only variant, \ie, EgoJEPA (Robot-only) in Figure~\ref{fig:mainRes}, which retains the same architecture but omits human-data training in Stages~1--2 and is trained directly on robot demonstrations. As shown, human pre-training increases the average out-of-domain success rate from 40\% to 74\%, with improvements across all five tasks. The largest gain occurs on \texttt{bottle cap unscrewing}, with success increasing from 30\% to 70\%. This task requires coordinated bimanual control to stabilize the bottle while rotating the cap, suggesting the value of human manipulation priors for learning coordinated actions. We further examine Stage~3 training dynamics on \texttt{bottle cap unscrewing} by comparing action loss over training steps between EgoJEPA and its robot-only variant. As shown in Figure~\ref{fig:training_data_efficiency} (left), EgoJEPA converges faster and achieves substantially lower training loss, suggesting that priors learned from egocentric human videos facilitate embodiment-specific adaptation under limited robot supervision.

\noindent\textbf{JEPA \emph{vs.} VQ-VAE.} To assess the benefit of JEPA-based predictive learning, we compare EgoJEPA with EgoJEPA (VQ-VAE), which replaces Stage~1 predictive learning with VQ-VAE-based action reconstruction~\citep{van2017neural}, using the same data and downstream training protocol. As shown in Figure~\ref{fig:mainRes}, EgoJEPA increases the average out-of-domain success rate from 63\% to 74\%, with gains across all five tasks. These results suggest that learning the visual consequences of human motion yields more effective representations for downstream control than reconstructing motion alone. Refer to Appendix \ref{app:implementation} for more details.

\input{table_figs/figRoboDataEfficiency}
\noindent\textbf{Robot Data Efficiency.} We evaluate data efficiency on \texttt{bottle cap unscrewing} using 10\%, 25\%, 50\%, and 100\% of the robot demonstration dataset during adaptation. As shown in Figure~\ref{fig:training_data_efficiency} (right), EgoJEPA consistently outperforms both robot-only training (\ie, EgoJEPA (Robot-only)) and the retargeting baseline across all data budgets. Notably, with only 10\% of the demonstrations, EgoJEPA achieves a 35\% success rate, surpassing the 30\% attained by the robot-only model trained on the full dataset (100\%). Similarly, using just 50\% of the demonstrations, EgoJEPA reaches a 65\% success rate, outperforming retargeting using all demonstrations (60\%). These results demonstrate that pre-trained effect-aware representations enable substantially more data-efficient policy adaptation, significantly reducing reliance on costly robot teleoperation data.

\input{table_figs/figRobustness}

\noindent\textbf{Robustness to Human Annotation Errors.} Hand-motion annotations in human videos are susceptible to localization errors, temporal misalignment, and tracking failures, which can propagate through retargeting into robot action supervision. To evaluate robustness against noisy human pre-training data, we independently introduce three types of perturbation to hand annotations during Stages~1--2: 1) \emph{Position noise}: additive Gaussian noise with per-axis standard deviation $\sigma \in \{0,2,5,10\}$~mm; 2) \emph{Temporal jitter}: random frame shifts relative to original video frames, sampled uniformly from $\{-\delta, \dots, \delta\}$ with maximum temporal offset $\delta \in \{0, 3, 6, 12\}$ frames; and 3) \emph{Annotation dropout}: masking contiguous annotation segments covering a fraction $p \in \{0\%, 10\%, 20\%, 40\%\}$ of total frames, with missing keypoints linearly interpolated. All variants are evaluated on \texttt{bottle cap unscrewing}, keeping Stage~3 robot demonstration data and training settings identical across experiments. See Appendix~\ref{app:robustness} for further implementation details.

As illustrated in Figure~\ref{fig:robustness}, EgoJEPA demonstrates significantly higher resilience to all three annotation error types compared with the retargeting pipeline. Under the maximum position noise ($\sigma = 10$~mm), EgoJEPA experiences a modest 10\% success-rate drop (70\% to 60\%), whereas retargeting suffers a drastic 30\% decline (60\% to 30\%). Among the evaluated perturbations, temporal jitter induces the most severe performance decay for both models, while annotation dropout exhibits the mildest impact. Specifically, at the highest perturbation levels, EgoJEPA's success rate degrades to 50\% under temporal jitter ($\delta = 12$) but maintains 65\% under annotation dropout ($p = 40\%$). 

\input{table_figs/figEffiencyFiveTasks}

\noindent\textbf{Cross-Embodiment Efficiency.} To further evaluate the computational efficiency, we record the wall-clock adaptation costs across five tasks and three distinct robot embodiments. Wall-clock time is reported rather than GPU hours because the human-robot calibration step in the retargeting baseline is executed entirely on CPUs, making real-time wall-clock duration a more consistent and fair measure of end-to-end deployment cost. As illustrated in Figure~\ref{fig:efficiencyLayout}, EgoJEPA seamlessly reuses pre-trained human action representations (Stages~1--2) across all downstream embodiments. In contrast, the retargeting pipeline requires re-running expensive calibration and latent world modeling whenever a new robot embodiment is introduced. For instance, when adapting to the Xense Tac UMI gripper for \texttt{cardboard box folding}, EgoJEPA requires only 37.5 hours of Stage~3 tuning, compared with 193.4 hours for retargeting (including 46.7 hours of CPU-based calibration and 109.2 hours of GPU-based world modeling). Across all five tasks, EgoJEPA reduces cumulative wall-clock training time from 743.2 hours to 392.7 hours, achieving a 47.2\% overall cost reduction. These results highlight EgoJEPA's substantial time and compute savings when transferring shared, embodiment-agnostic action priors to new robotic platforms.

%% file: table_figs/figMainRes.tex
\begin{figure*}[t]
	\centering
    \includegraphics[width=\textwidth]{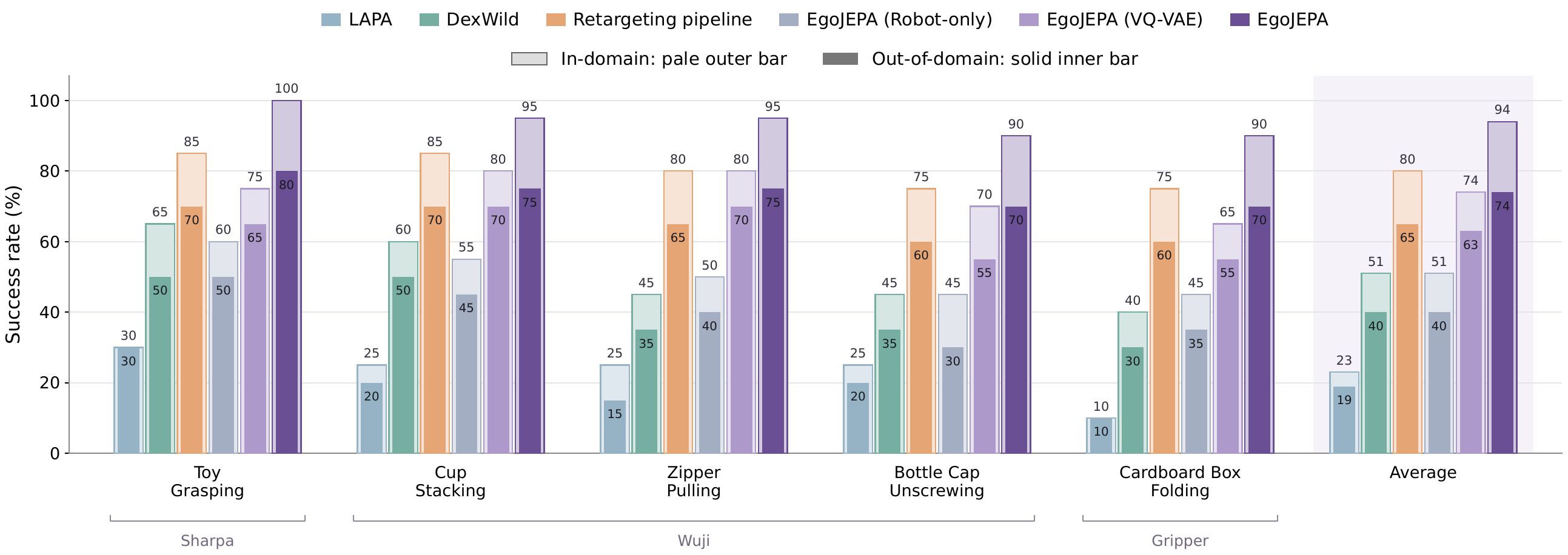}
    \caption{\textbf{In-domain and out-of-domain manipulation performance} across five tasks and three embodiments including Sharpa, Wuji, and Gripper. Bars report success rates over 20 trials per task, with the average success rate across tasks shown on the right.}
	\label{fig:mainRes}
\end{figure*}

%% file: table_figs/figVisMainResults.tex
\begin{figure*}[t]
	\centering
    \includegraphics[width=\textwidth]{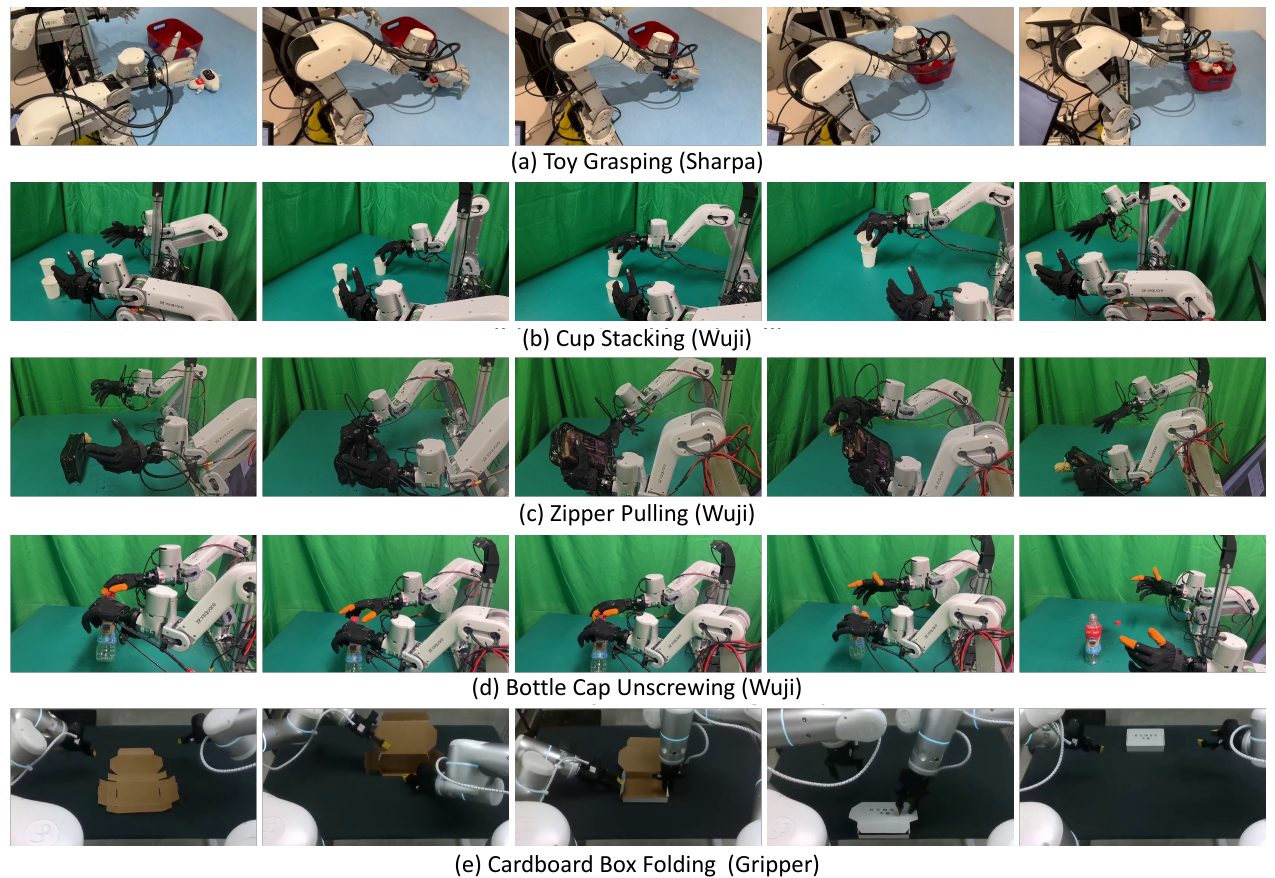}
    \caption{\textbf{Qualitative results} on five real-world robotic manipulation tasks.}
	\label{fig:visMainRes}
\end{figure*}

%% file: table_figs/figRoboDataEfficiency.tex
\begin{figure}[t]
    \centering
    \begin{subfigure}[t]{0.43\linewidth}
        \centering
        \includegraphics[width=0.96\linewidth]{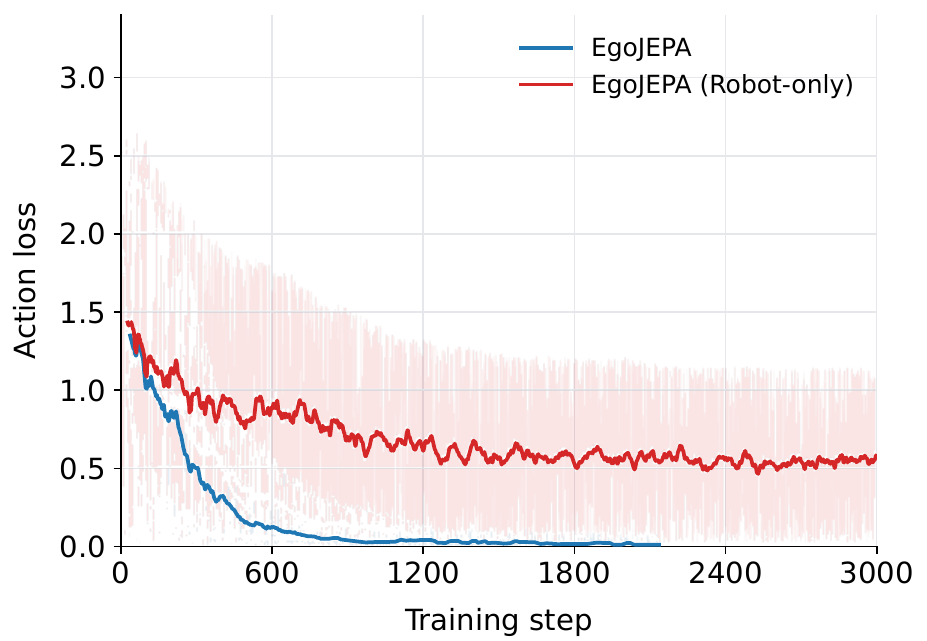}
        \label{fig:training_convergence}
    \end{subfigure}
    \hspace{2mm}
    \begin{subfigure}[t]{0.43\linewidth}
        \centering
        \includegraphics[width=0.92\linewidth]{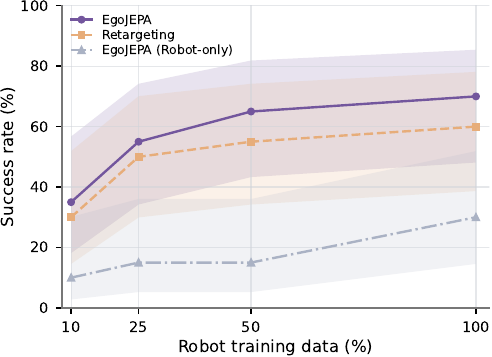}
        \label{fig:robot_data_efficiency}
    \end{subfigure}
    \vspace{-2mm}
    \caption{\textbf{Left: Training convergence comparisons.} Action loss during Stage~3 training on robot data for EgoJEPA and its robot-only variant. \textbf{Right: Robot data efficiency analysis.} Success rates on the task of \texttt{bottle cap unscrewing} with varying fractions of robot demonstrations. Shaded regions indicate 95\% Wilson confidence intervals based on 20 evaluation trials per setting.}
    \vspace{-3mm}
    \label{fig:training_data_efficiency}
\end{figure}

%% file: table_figs/figRobustness.tex
\begin{figure*}[t]
	\centering
\includegraphics[width=0.96\textwidth]{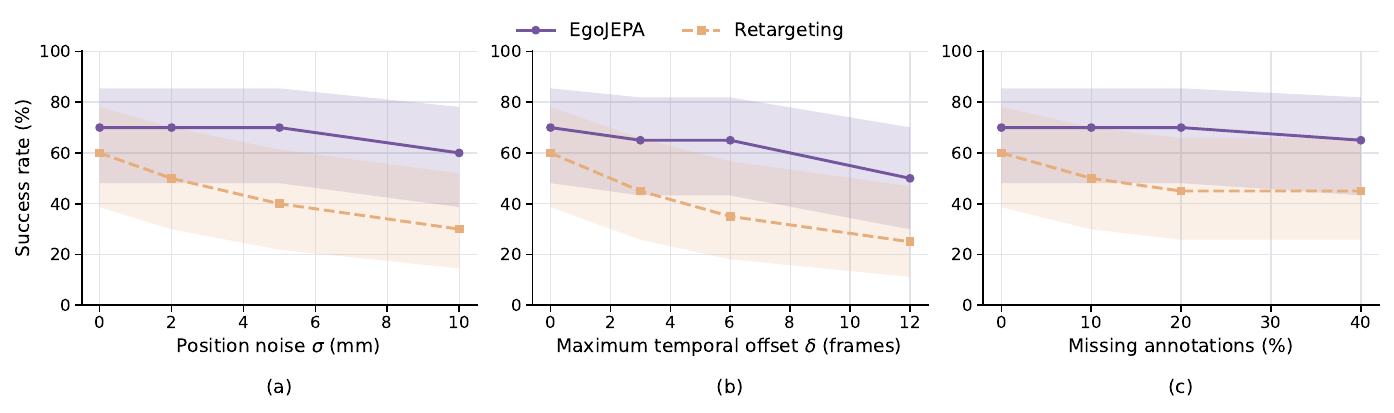}
\vspace{-4mm}
\caption{\textbf{Robustness analysis to human action annotation errors.} The success rate on \texttt{bottle cap unscrewing} with increasing (a) position noise standard deviation $\sigma$, (b) maximum temporal offset $\delta$ and (c) missing-annotation fraction $p$ in human pre-training data (stage 1 \& 2).}
\vspace{-3mm}
\label{fig:robustness}
\end{figure*}

%% file: table_figs/figEffiencyFiveTasks.tex
\begin{figure*}[t]
	\centering
    \includegraphics[width=0.93\textwidth]{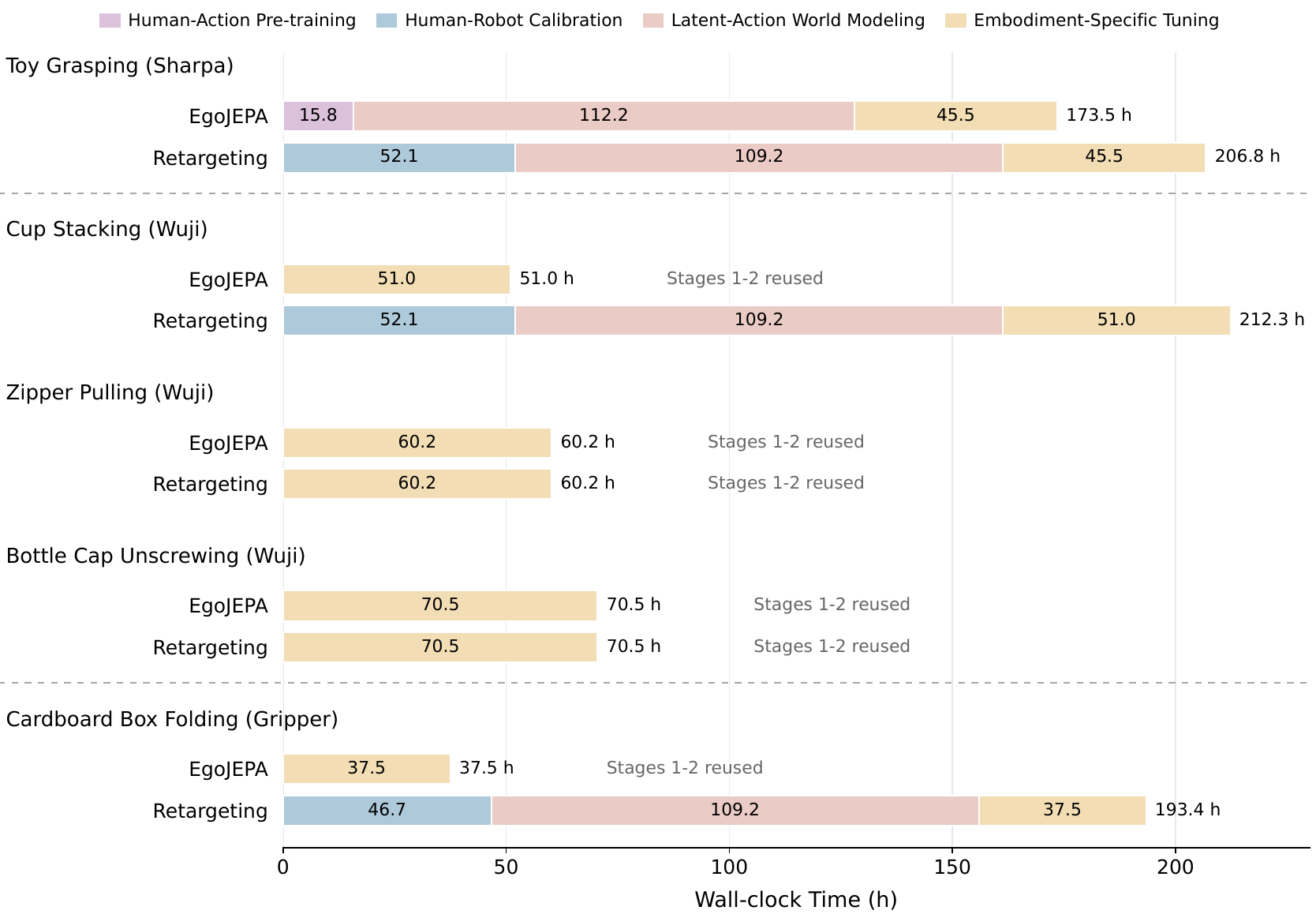}
    \caption{\textbf{Training efficiency across five tasks and three robot embodiments.} Stacked bars show the wall-clock time required for each task, with colors denoting training stages. EgoJEPA achieves greater training efficiency by reusing human pre-training (Stages~1--2) across robot embodiments.}
	\label{fig:efficiencyLayout}
	\vspace{-2mm}
\end{figure*}

%% file: sec/5_con.tex
\section{Conclusion}

We introduced EgoJEPA, a three-stage framework that transfers manipulation priors from egocentric human videos to robot control without rule-based retargeting. By grounding latent actions in visual effects and integrating them into world modeling, EgoJEPA decouples human pre-training from embodiment-specific control, enabling shared pretrained models to be reused across robots through lightweight action adapters. Experiments across five real-world tasks and three embodiments demonstrate that EgoJEPA outperforms the retargeting pipeline by an absolute margin of 9\% under out-of-domain evaluation while reducing cumulative cross-embodiment training time by 47.2\%. In addition, EgoJEPA exhibits greater robustness to human annotation noise, supporting scalable learning from imperfectly annotated egocentric videos.

%% file: sec/6_statement.tex
\subsection*{AI use statement}
In this work, we used generative AI tools to assist with literature searches, manuscript polishing, and experimental design. We have not used generative AI tools to generate synthetic datasets or implement the proposed method. Developing theoretical models, formulating mathematical claims, and writing proofs are not applicable to this work. We reviewed and revised the AI-assisted contents against experimental records, and verified references against their original sources. We take responsibility for the final content of this work, including text, claims, and artifacts produced with the aid of generative AI.

%% file: sec/6_appendix.tex
\section{Appendix}

\setcounter{equation}{0}
\setcounter{figure}{0}
\setcounter{table}{0}

\renewcommand{\theequation}{A\arabic{equation}}
\renewcommand{\thefigure}{A\arabic{figure}}
\renewcommand{\thetable}{A\arabic{table}}

\subsection{Dataset Details} \label{app:datasets}

\input{table_figs/tabStatisticPretrain}

\noindent\textbf{Human Pre-training Data Statistics.} Table~\ref{tab:pretrain_statistics} summarizes the dataset scale, recording specifications, and annotation coverage. Figure~\ref{fig:pretrainTaskScene} presents the 20 most frequent tasks and scenes, covering household activities and industrial operations such as clothes folding, dish washing, and component assembly. Figure~\ref{fig:pretrainLength} shows the range of episode durations and frame counts, encompassing both brief interactions and extended manipulation sequences. Figure~\ref{fig:visHumanHand} further illustrates representative egocentric video sequences with aligned hand-keypoint annotations, providing a qualitative view of the human pre-training data.

\input{table_figs/figPretrainTask}

\input{table_figs/figPretrainLength}

\input{table_figs/figVisHumanHand}


\noindent\textbf{Robot Demonstration Statistics.} Table~\ref{tab:evaluation_tasks} summarizes the embodiment, action dimensionality, episode count, and average duration for each task. The datasets cover five manipulation tasks using the Sharpa hand, Wuji hand, and Xense Tac UMI gripper, with action dimensions of 56, 52, and 14, respectively. Demonstration durations vary with task demands, ranging from brief grasping interactions to longer sequences involving coordinated manipulation.

\input{table_figs/tabStatisticPosttrain}

\input{table_figs/tabTrainConfig}

\noindent\textbf{Data Licensing and Privacy Safeguards.} To ensure strict regulatory compliance and ethical standards, all third-party and commercially purchased datasets utilized in our experiments are managed under rigorous licensing and privacy protocols:
\begin{itemize}[leftmargin=*, noitemsep, topsep=2pt]
    \item \emph{Commercial Licensing \& Usage Scope:} All commercially acquired datasets were legally licensed from standard commercial vendors under explicit agreements that grant authorization for non-commercial academic research and algorithmic evaluation. The usage fully complies with the intellectual property rights and End User License Agreements (EULA) specified by the vendors.
    \item \emph{Privacy Safeguards \& Anonymization:} Prior to model training and representation learning, all datasets underwent thorough privacy verification. Human subjects appearing in egocentric videos and annotations are strictly anonymized, with personally identifiable information (PII), \eg facial features, license plates, and sensitive background information, either blurred or removed by the data providers in accordance with global privacy regulations (\eg, GDPR).
    \item \emph{Data Security \& Responsible AI:} All purchased assets are securely stored on restricted-access servers with role-based access control to prevent unauthorized redistribution or leakages. No biometric identifiers or sensitive personal characteristics were collected, inferred, or processed.
\end{itemize}

\subsection{Implementation Details} \label{app:implementation}

\noindent\textbf{Training Configurations of EgoJEPA.}  Table~\ref{tab:training_config} summarizes the data, model initialization, and optimization settings for all three stages. Stages~1--2 use egocentric human data, while Stage~3 initializes from the Stage~2 checkpoint and jointly tunes the video and action Transformers with embodiment-specific adapters on robot data. Stage~3 additionally incorporates three camera views and proprioceptive conditioning. All stages use AdamW, with stage-specific learning rates, schedules, and batch sizes as detailed in the table.

\textbf{Implementations of EgoJEPA (VQ-VAE).} We detail the implementation and training procedure of EgoJEPA (VQ-VAE), which replaces Stage~1 JEPA pre-training with VQ-VAE-based hand-motion reconstruction \citep{van2017neural}. Let $\mathcal{D}_{\mathrm{ego}}$ denote an egocentric human video dataset, where each trajectory contains RGB observations $\{o_t\}_{t=1}^{T}$ and aligned hand annotations $\{m_t\}_{t=1}^{T}$. A VQ-VAE encoder $f_{\mathrm{enc}}$ maps each hand annotation $m_t$ to a continuous representation $\mathbf{h}_t$, which is quantized using a
learnable codebook $\mathcal{C}=\{\mathbf{e}_k\}_{k=1}^{K}$ as follows:
\begin{equation}
    \mathbf{q}_t
    = \mathbf{h}_t + \operatorname{sg}(\mathbf{e}_{k_t}-\mathbf{h}_t),
    \quad \emph{s.t.} \quad
    k_t = \arg\min_{k\in\{1,\ldots,K\}}
    \|\mathbf{h}_t-\mathbf{e}_k\|_2^2,
\end{equation}
where $\operatorname{sg}(\cdot)$ is a straight-through estimator for encoder gradients, which preserves its input during the forward pass and blocks gradients during backpropagation.

Then, a VQ-VAE decoder $f_{\mathrm{dec}}$ maps the quantized representation $\mathbf{q}_t$ back to the hand annotation space, yielding $\hat{m}_t=f_{\mathrm{dec}}(\mathbf{q}_t)$. The overall training objective is as follows:
\begin{equation}
    \mathcal{L}_{\mathrm{VQ}}
    =
    \mathbb{E}_{\tau\sim\mathcal{D}_{\mathrm{ego}},\,t}
    \left[
        \underbrace{\|m_t-\hat{m}_t\|_2^2}_{\mathcal{L}_{\mathrm{rec}}}
        +
        \underbrace{
            \|\operatorname{sg}(\mathbf{h}_t)-\mathbf{e}_{k_t}\|_2^2
        }_{\mathcal{L}_{\mathrm{codebook}}}
        +
        \underbrace{
            \|\mathbf{h}_t-\operatorname{sg}(\mathbf{e}_{k_t})\|_2^2
        }_{\mathcal{L}_{\mathrm{commit}}}
    \right],
\end{equation}
This VQ-VAE objective learns discrete representations of per-frame hand annotations without visual-effect supervision. It replaces only Stage~1 human-action pre-training, while the training protocols for Stages~2 and~3 remain unchanged.

\noindent\textbf{Implementations of Retargeting Pipeline.} We convert human wrist poses and 3D hand keypoints into embodiment-specific robot action targets. For the Sharpa and Wuji hands, we use their respective official retargeting implementations\footnote{\url{https://github.com/sharpa-robotics/sharpa-vision-retargeting-sdk}}\footnote{\url{https://github.com/wuji-technology/wuji-retargeting}}, adapting the input keypoint ordering and coordinate conventions to each interface. Wrist poses are mapped to robot end-effector poses through calibrated coordinate transformations and tool offsets. For the Xense Tac UMI gripper, the thumb--index fingertip distance is mapped to the gripper opening using a calibrated, range-constrained mapping. Retargeted trajectories are aligned with the video frames and converted into relative actions consistent with the robot demonstrations.

\subsection{Robustness Evaluation Protocol} \label{app:robustness}

We detail the experimental settings for evaluating robustness to human annotation errors. We independently introduce position noise, temporal jitter, and annotation dropout into human pre-training annotations, using identical perturbations for EgoJEPA and the retargeting pipeline. Stage~3 training remains unchanged, and both methods are evaluated on \texttt{bottle cap unscrewing} under the out-of-domain setting.

\noindent \emph{1) Position Noise.}
To evaluate robustness to localization errors in human hand-motion annotations, we perturb the annotated 3D positions with additive Gaussian noise. For each frame $t$, a translation vector is sampled independently and applied consistently to all keypoints of that hand:
\begin{equation}
    \widetilde{\mathbf{p}}_{t,j}
    = \mathbf{p}_{t,j} + \boldsymbol{\epsilon}_{t},
    \qquad
    \boldsymbol{\epsilon}_{t}
    \sim \mathcal{N}\!\left(\mathbf{0}, \sigma^2\mathbf{I}_3\right),
\end{equation}
where $\mathbf{p}_{t,j}\in\mathbb{R}^3$ denotes the position of keypoint $j$, and $\sigma$ is the noise standard deviation along each coordinate axis. The same translation is applied to the wrist position, preserving the hand geometry and leaving rotation annotations unchanged. We set $\sigma\in\{0,2,5,10\}$~mm, with $\sigma=0$ denoting clean annotations. Perturbations are introduced before computing inter-frame motion differences or performing retargeting. Both methods are pretrained using the same perturbed annotations and subsequently adapted and evaluated using clean robot data.

\noindent \emph{2) Temporal Jitter.} To evaluate robustness to temporal misalignment between human videos and hand-motion annotations, we randomly shift the annotation sequence relative to the video. For each training clip, an integer offset $\delta$ is sampled uniformly:
\begin{equation}
    \widetilde{\mathbf{m}}_t = \mathbf{m}_{t+\delta},
    \qquad
    \delta \sim \operatorname{Uniform}\{-k,\ldots,k\},
\end{equation}
where $\mathbf{m}_t$ denotes the hand-motion annotation at frame $t$, and $k$ controls the maximum temporal offset. The offset is shared across both hands and held constant within each clip, preserving motion continuity while disrupting visual--motion alignment. We set $k\in\{0,3,6,12\}$ frames, corresponding to maximum offsets of approximately $\{0,100,200,400\}$~ms at 30~Hz. Perturbations are applied before temporal subsampling and action extraction, using only clips with sufficient surrounding frames to accommodate the shifts. Both methods use the same perturbed annotations for human pre-training, followed by adaptation and evaluation on clean robot data.

\noindent \emph{3) Annotation Dropout.}
To evaluate robustness to missing hand-motion annotations caused by occlusion or tracking failures, we randomly mask contiguous annotation segments within each training clip. Let $\mathcal{D}$ denote the set of masked frames. The perturbed annotations are obtained as
\begin{equation}
    \widetilde{\mathbf{m}}_t =
    \begin{cases}
        \mathbf{m}_t, & t \notin \mathcal{D}, \\
        \operatorname{Interp}(\mathbf{m}_{t^-},\mathbf{m}_{t^+};t),
        & t \in \mathcal{D},
    \end{cases}
\end{equation}
where $t^-$ and $t^+$ are the nearest valid frames before and after the missing segment. Positions are reconstructed by linear interpolation, and rotations by spherical linear interpolation. 
Both the retargeting pipeline and our EgoJPEA use identical masks and interpolation rules for human pre-training, followed by adaptation and evaluation on clean robot data.

\subsection{Broader Impact and Limitations}

\noindent\textbf{Broader Impact.} This work advances generalist robot learning by presenting an embodiment-agnostic framework that effectively leverages large-scale, low-cost human egocentric videos. By bypassing tedious manual geometric retargeting and repeated human-robot calibrations, EgoJEPA significantly lowers the computational and labor barriers to deploying multi-embodiment dexterous robotic systems in real-world environments, such as domestic assistance and industrial automation. However, scaling learning from uncurated video datasets carries the risk of inheriting or amplifying biases present in human demonstration environments. Furthermore, as autonomous dexterous manipulation systems become more capable in physical deployment, safety, alignment, and strict physical containment protocols must be rigorously established to prevent unintended hardware collisions or operational hazards.

\noindent\textbf{Limitations.} Although EgoJEPA demonstrates strong generalization and cross-embodiment efficiency for dexterous manipulation, several limitations remain: \emph{(1) Performance bounds under severe action quality issues:} While EgoJEPA exhibits noise robustness, extreme degradation in egocentric video actions (\eg, severe temporal misalignments or prolonged occlusions) can still distort latent representations and bottleneck downstream task performance. \emph{(2) Contact-intensive dynamics:} Performance drops on out-of-domain tasks that require fine-grained contact dynamics (\eg, \texttt{bottle cap unscrewing}). Integrating explicit interaction modeling or force-feedback priors into the latent action space remains a key direction for future work.

%% file: table_figs/tabStatisticPretrain.tex
\begin{table}[t]
    \centering
    \caption{\textbf{Summary statistics of the human pre-training dataset.} Dataset scale, recording specifications, task and scene coverage, and hand annotations.}
    \label{tab:pretrain_statistics}
    \small
    \setlength{\tabcolsep}{8pt}
    \renewcommand{\arraystretch}{1.1}
    \begin{tabular}{lr}
        \toprule
        \textbf{Statistic} & \textbf{Value} \\
        \midrule
        Episodes & 1,111,943 \\
        Video Frames & 2,169,344,852 \\
        Total Duration & ${\approx}20,106.6$ hours \\
        Frame Rate & ${\approx}29.97$ Hz \\
        Camera Views & 1 \\
        Mean Episode Duration & ${\approx}65$ s \\
        Median Episode Duration & ${\approx}30$ s \\
        Distinct Task Instructions & 64,686 \\
        Scene Labels & 608 \\
        3D Keypoints per Hand & 21 \\
        \bottomrule
    \end{tabular}
\end{table}

%% file: table_figs/figPretrainTask.tex
\begin{figure*}[t]
	\centering
    \includegraphics[width=\textwidth]{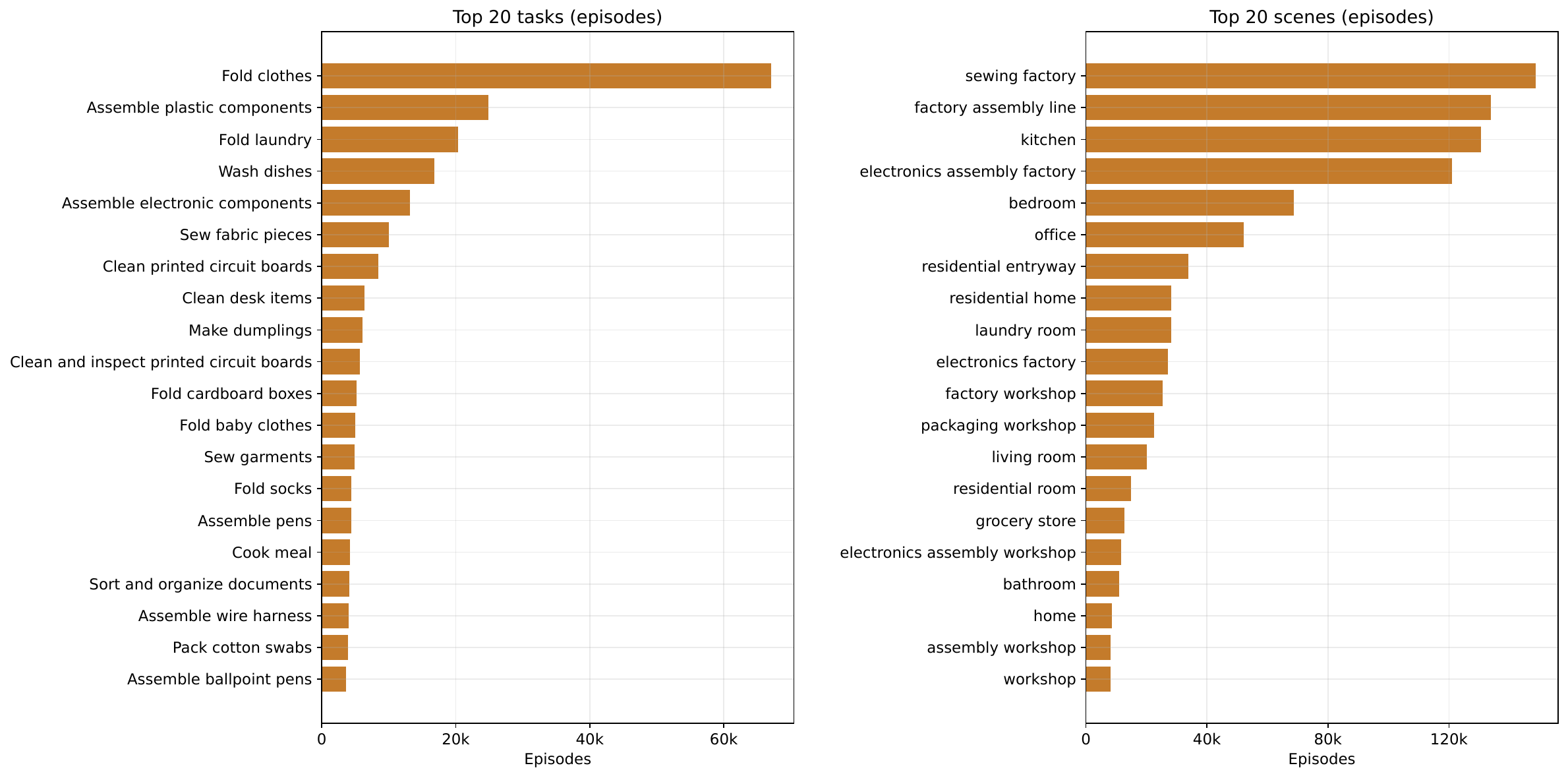}
    \caption{\textbf{Task and scene distributions of the human pre-training dataset.} The 20 most frequent tasks (left) and scenes (right), ranked by episode count.}
	\label{fig:pretrainTaskScene}
\end{figure*}

%% file: table_figs/figPretrainLength.tex
\begin{figure*}[t]
	\centering
    \includegraphics[width=\textwidth]{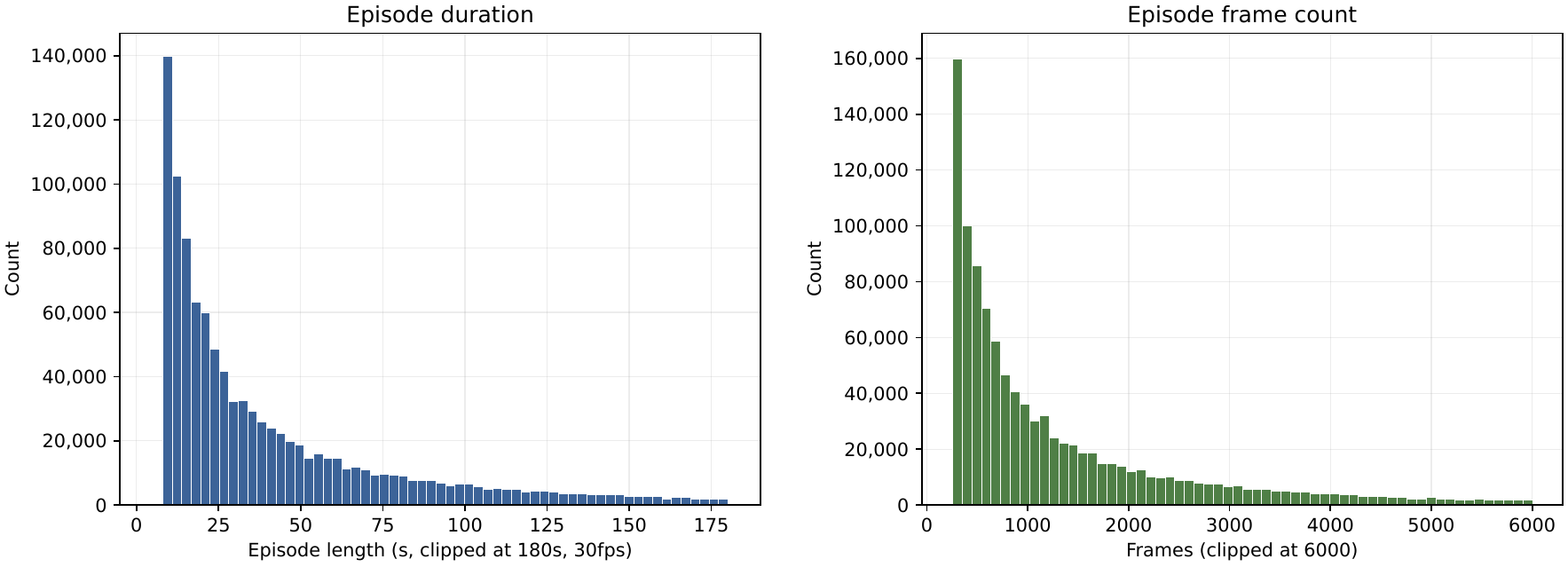}
    \caption{\textbf{Episode length distributions of the human pre-training dataset.} Histograms show episode durations in seconds (left) and frame counts (right).}
	\label{fig:pretrainLength}
\end{figure*}

%% file: table_figs/figVisHumanHand.tex
\begin{figure*}[t]
	\centering
        \includegraphics[width=0.96\textwidth]{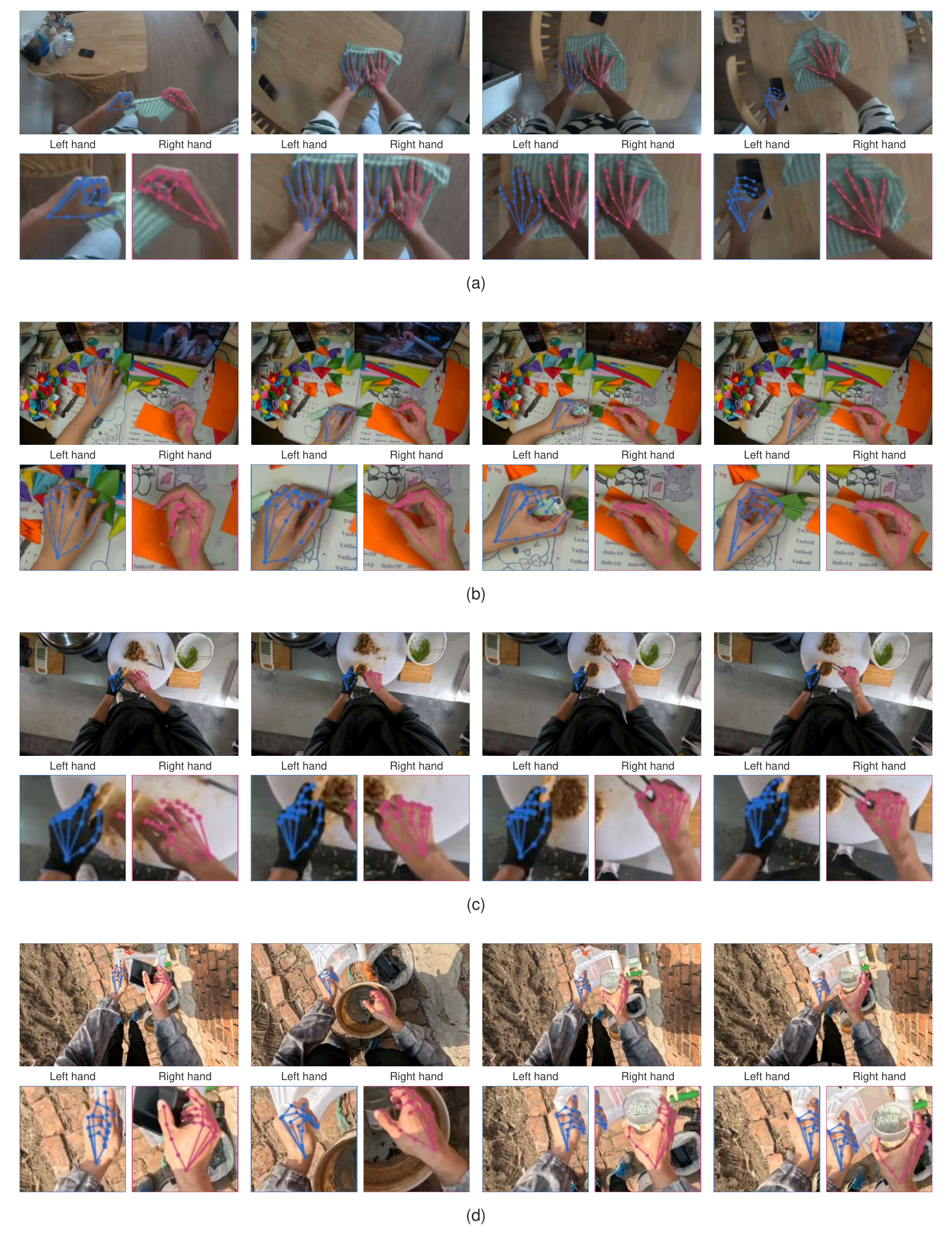}
        \vspace{-2mm}
        \caption{\textbf{Examples of egocentric human videos and hand annotations.} Each sequence shows sampled frames with enlarged hand crops, where blue and pink keypoints denote the left and right hands, respectively.}
	\label{fig:visHumanHand}
\end{figure*}

%% file: table_figs/tabStatisticPosttrain.tex
\begin{table*}[t]
    \centering
    \caption{\textbf{Task-wise statistics} of the robot demonstration datasets.}
    \label{tab:evaluation_tasks}
    \small
    \setlength{\tabcolsep}{10pt}
    \renewcommand{\arraystretch}{1.1}
    \resizebox{\textwidth}{!}{
    \begin{tabular}{lccccc}
        \toprule
        \textbf{Task} & \textbf{Embodiment}
        & \textbf{Action Dim.}
        & \textbf{\# Episode}
        & \textbf{Avg. Length (s)} \\
        \midrule
        Plush Toy Grasping    & Sharpa Hand & 56 & 750 & 6.9  \\
        Cup Stacking          & Wuji Hand   & 52 & 916 & 13.6 \\
        Zipper Manipulation   & Wuji Hand   & 52 & 623 & 15.5  \\
        Bottle Cap Unscrewing & Wuji Hand   & 52 & 635 & 23.8 \\
        Paper-Box Folding     & Xense Tac UMI gripper & 14 & 843 & 297.6 \\
        \bottomrule
    \end{tabular}
    }
\end{table*}


%% file: table_figs/tabTrainConfig.tex
\begin{table*}[t]
    \centering
    \caption{\textbf{Training configurations} for the three stages of EgoJEPA.}
    \label{tab:training_config}
    \setlength{\tabcolsep}{8pt}
    \renewcommand{\arraystretch}{1.12}
    \resizebox{\textwidth}{!}{
    \begin{tabular}{lccc}
        \toprule
        \textbf{Configuration}
        & \textbf{Stage~1}
        & \textbf{Stage~2}
        & \textbf{Stage~3} \\
        \midrule
        Training Data
        & Ego Data
        & Ego Data
        & Robot Data \\

        Initialization
        & Wan2.2 VAE
        & Wan2.2-TI2V-5B
        & Stage~2 Checkpoint \\

        Trainable Modules
        & All Modules
        & Video DiT + Action DiT
        & Video DiT + Action DiT + Adapters \\

        Input Resolution
        & $256\times256$
        & $192\times224$
        & $192\times224$ \\

        Camera Views
        & 1 & 1 & 3 \\

        Proprioceptive Conditioning
        & No & No & Yes \\

        Action Prediction Horizon
        & -- & 32 & 32 \\

        Latent Action Dimension
        & 1024 & 128 & 128 \\

        Robot Data Dimension
        & -- & -- & 56/52/14 \\
        
        Number of GPUs 
        & 32 & 64 & 64 \\

        Per-device Batch Size
        & 128 & 128 & 64 \\

        Steps 
        & 3,500 & 6,000 & 3,000 \\

        Optimizer
        & AdamW & AdamW & AdamW \\

        AdamW $\boldsymbol{\beta}$
        & $(0.9, 0.999)$ & $(0.95, 0.999)$ & $(0.95, 0.999)$ \\

        Learning Rate
        & $4.25\times10^{-4}$
        & $3\times10^{-5}$
        & $1\times10^{-4}$ \\

        Minimum Learning Rate
        & 0
        & $1\times10^{-6}$
        & $1\times10^{-6}$ \\

        LR Schedule
        & Linear & Cosine & Cosine \\


        Weight Decay
        & 0.04 & 0.01 & 0.01 \\

        Maximum Gradient Norm
        & -- & 1.0 & 1.0 \\



        \bottomrule
    \end{tabular}
    }
\end{table*}